\documentclass[conference]{IEEEtran}
\IEEEoverridecommandlockouts
\usepackage{cite}
\usepackage{amsmath,amssymb,amsfonts}
\usepackage{graphicx}
\usepackage{textcomp}
\usepackage{xcolor}

\usepackage[linesnumbered,ruled,vlined]{algorithm2e}%[ruled,vlined]{
\usepackage{amsmath}
\usepackage{subcaption}
\usepackage{booktabs}
\usepackage{pifont}
\usepackage{multirow}
\usepackage{tabularray}
\usepackage{algpseudocode}  
\DeclareMathOperator*{\argmin}{arg\,min}

\makeatletter
\newcommand{\linebreakand}{%
  \end{@IEEEauthorhalign}
  \hfill\mbox{}\par
  \mbox{}\hfill\begin{@IEEEauthorhalign}
}
\makeatother

\def\BibTeX{{\rm B\kern-.05em{\sc i\kern-.025em b}\kern-.08em
    T\kern-.1667em\lower.7ex\hbox{E}\kern-.125emX}}

\renewcommand{\baselinestretch}{0.995}
    
\begin{document}

\title{FedAH: Aggregated Head for Personalized Federated Learning}

\author{\IEEEauthorblockN{1\textsuperscript{st} Pengzhan Zhou\thanks{*Pengzhan Zhou is the corresponding author.}\IEEEauthorrefmark{1}}
\IEEEauthorblockA{\textit{College of Computer Science} \\
\textit{Chongqing University}\\
Chongqing, China \\
pzzhou@cqu.edu.cn}
\and
\IEEEauthorblockN{2\textsuperscript{nd} Yuepeng He}
\IEEEauthorblockA{\textit{College of Computer Science} \\
\textit{Chongqing University}\\
Chongqing, China \\
 hyp@stu.cqu.edu.cn}
\and
\IEEEauthorblockN{3\textsuperscript{rd} Yijun Zhai}
\IEEEauthorblockA{\textit{College of Computer Science} \\
\textit{Chongqing University}\\
Chongqing, China \\
yjzhai@stu.cqu.edu.cn}
\and
\IEEEauthorblockN{4\textsuperscript{th} Kaixin Gao}
\IEEEauthorblockA{\textit{College of Computer Science} \\
\textit{Chongqing University}\\
Chongqing, China \\
kaixingaocs@stu.cqu.edu.cn}
\linebreakand
\IEEEauthorblockN{5\textsuperscript{th} Chao Chen}
\IEEEauthorblockA{\textit{College of Computer Science} \\
\textit{Chongqing University}\\
Chongqing, China \\
cschaochen@cqu.edu.cn}
\and
\IEEEauthorblockN{6\textsuperscript{th} Zhida Qin}
\IEEEauthorblockA{\textit{School of Computer Science and Technology} \\
\textit{Beijing Institute of Technology}\\
Beijing, China\\
zanderqin@bit.edu.cn}
\and
\IEEEauthorblockN{7\textsuperscript{th} Chong Zhang}
\IEEEauthorblockA{\textit{School of Computer Science and Technology} \\
\textit{Xi'an Jiaotong University}\\
Xi'an, China \\
zhangchong@xjtu.edu.cn}
\linebreakand
\IEEEauthorblockN{8\textsuperscript{th} Songtao Guo}
\IEEEauthorblockA{\textit{College of Computer Science} \\
\textit{Chongqing University}\\
Chongqing, China \\
guosongtao@cqu.edu.cn}
}

\maketitle

\begin{abstract}
Recently, Federated Learning (FL) has gained popularity for its privacy-preserving and collaborative learning capabilities. Personalized Federated Learning (PFL), building upon FL, aims to address the issue of statistical heterogeneity and achieve personalization. Personalized-head-based PFL is a common and effective PFL method that splits the model into a feature extractor and a head, where the feature extractor is collaboratively trained and shared, while the head is locally trained and not shared. However, retaining the head locally, although achieving personalization, prevents the model from learning global knowledge in the head, thus affecting the performance of the personalized model. To solve this problem, we propose a novel PFL method called Federated Learning with Aggregated Head (FedAH), which initializes the head with an Aggregated Head at each iteration. The key feature of FedAH is to perform element-level aggregation between the local model head and the global model head to introduce global information from the global model head. To evaluate the effectiveness of FedAH, we conduct extensive experiments on five benchmark datasets in the fields of computer vision and natural language processing. FedAH outperforms ten state-of-the-art FL methods in terms of test accuracy by 2.87\%. Additionally, FedAH maintains its advantage even in scenarios where some clients drop out unexpectedly. Our code is open-accessed at https://github.com/heyuepeng/FedAH.

\end{abstract}


\section{Introduction}

The traditional centralized training approach in machine learning is facing significant challenges due to the increasing importance of user data privacy \cite{yang2020federated, nguyen2021federated, zhang2023fedcp}. On the other hand, it is difficult to achieve well-performing models on individual clients due to the sparsity of data on each client \cite{kairouz2021advances,li2020federated, tan2022towards}. Federated Learning (FL), as a popular distributed machine learning paradigm, offers excellent privacy protection and collaborative learning capabilities \cite{zhang2024eliminating}. The learning tasks in FL are coordinated by a server and solved collaboratively by a network of multiple participating devices (clients). FedAvg \cite{mcmahan2017communication} is the original FL method and serves as the fundamental framework for subsequent FL methods. Its iterative process can be described in five steps: (1) The server randomly selects a subset of clients to join FL and distributes the same global model to them for initialization; (2) Clients overwrite their local model parameters with the parameters of the downloaded global model to acquire global knowledge; (3) Clients train their local models on their private local data; (4) Clients upload their trained local models to the server; (5) The server receives the local models from clients and performs weighted averaging on the model parameters to obtain a new global model. FedAvg aims to learn a single global model that performs well across all clients. However, this approach often suffers in statistically heterogeneous environments, such as when facing not independent and identically distributed (Non-IID) and unbalanced data \cite{li2020federated, yang2019federated}, leading to degraded model performance \cite{kairouz2021advances, huang2021personalized, t2020personalized}.

Personalized Federated Learning (PFL) has been proposed to address statistical heterogeneity and achieve personalization in FL \cite{tan2022towards}. PFL focuses on learning personalized models rather than a single global model \cite{t2020personalized}. For each client participating in FL, the global model distributed by the server contains global information, which can enhance the local model and address the data scarcity issue of clients. Most existing PFL methods use the global model as a container for global information and exploit global/personalized information by leveraging the parameters of the global/local models \cite{zhang2023fedcp}. Specifically, meta-learning-based PFL methods (such as Per-FedAvg \cite{fallah2020personalized} and FedMeta \cite{chen2018federated}) adapt the global model parameters to heterogeneous client data through fine-tuning. Regularization-based PFL methods (such as FedProx \cite{li2020federated}, pFedMe \cite{t2020personalized}, and Ditto \cite{li2021ditto}) regularize the model parameters during local training. Personalized-aggregation-based PFL methods (such as FedFomo \cite{zhang2020personalized}, APPLE \cite{luo2022adapt}, FedAMP \cite{huang2021personalized}, and FedALA \cite{zhang2023fedala}) achieve better local initialization by aggregating the models of other clients or combining global and local models. Personalized-head-based PFL methods (such as FedPer \cite{arivazhagan2019federated} and FedRep \cite{collins2021exploiting}) split the model into a global part (feature extractor) and a personalized part (head), with the feature extractor trained collaboratively and shared among clients, while the head is trained locally and not shared. This approach aims to utilize both global and personalized information in the model parameters. However, training the head only with local data can result in the loss of some global information in the head, negatively impacting the performance of the personalized model.

\begin{figure}[htpb]
    \centering
    \includegraphics[width=\linewidth]{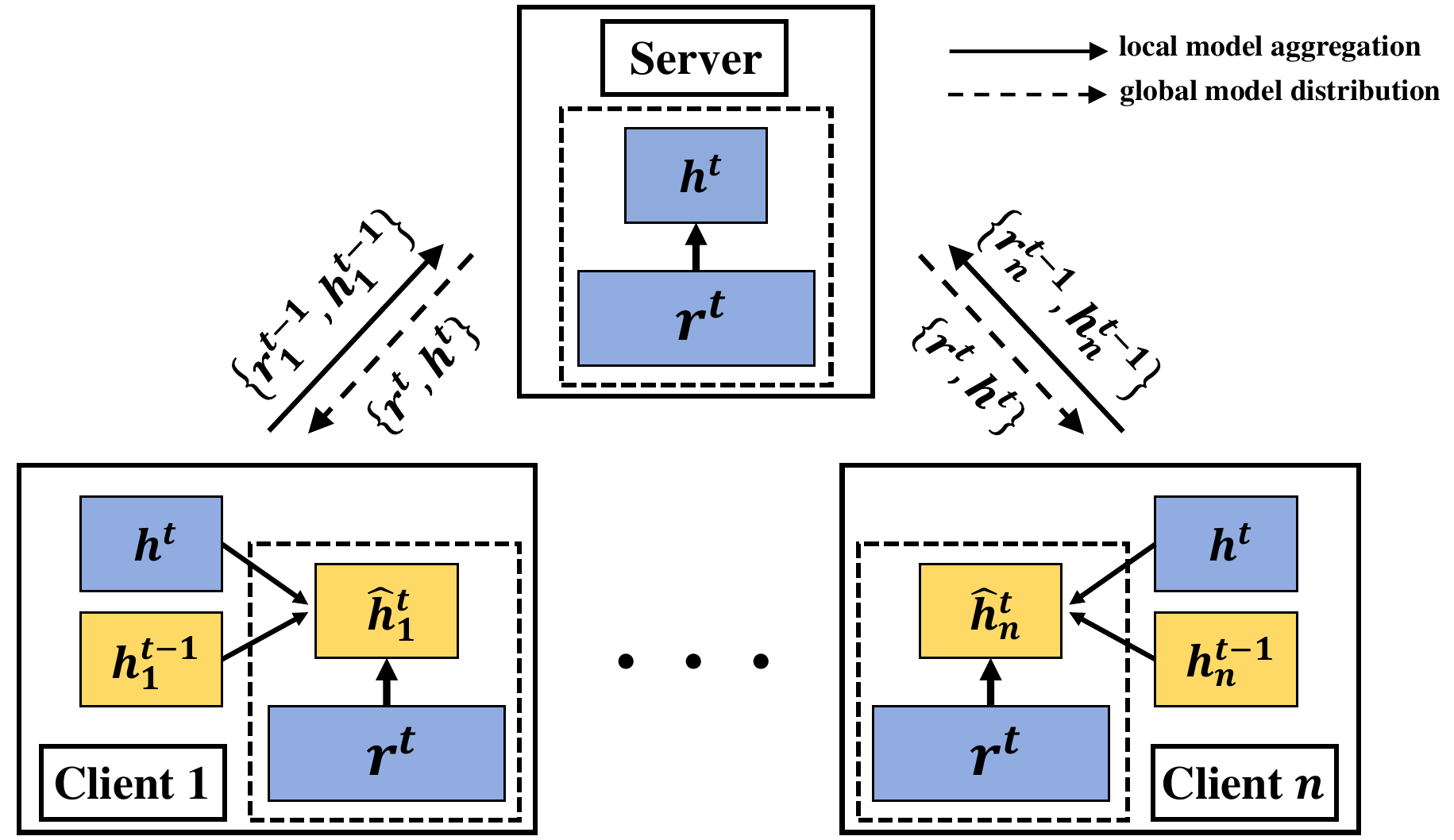}
    \caption{An example for FedAH. $r^t$ : global feature extractor; $h^t$ : global heads; $h^{t-1}_{1},\ldots,h^{t-1}_{n}$: local head of clients $1,\ldots, n$  in the previous iteration; $\hat h^t_{1},\ldots,\hat h^t_{n}$ : Aggregated Heads of clients $1,\ldots, n$  in the current iteration.}
    \label{fig:fedah}
\end{figure}

To address the issue of losing global information in the personalized head of personalized-head-based PFL methods, we propose a novel PFL method, Federated Learning with Aggregated Head (FedAH). As shown in Figure \ref{fig:fedah}, FedAH combines the ideas of personalized-aggregation-based PFL by aggregating the local model head from the previous iteration and the global model head from the current iteration at the element level to obtain an Aggregated Head, thereby integrating global information into the global model head. Apart from using the Aggregated Head as the initialization head for a new iteration, the rest of the processes are motivated by FedRep, and the aggregation weights of the Aggregated Head are learned through gradient descent. By combining the aforementioned strategies, FedAH can achieve personalization while more comprehensively benefiting from global knowledge, thereby improving the performance of personalized models.

To evaluate the effectiveness of FedAH, we conduct extensive experiments in two widely adopted scenarios \cite{mcmahan2017communication, li2021model} (i.e., the pathological and practical heterogeneous settings) and five benchmark datasets. The experimental results demonstrate that FedAH outperforms ten state-of-the-art (SOTA) FL methods. In summary, our contributions are mainly three-fold:

\begin{itemize}
\item To the best of our knowledge, we are the first to consider introducing global information through personalized aggregation in the model head. This approach is more fine-grained and effective compared to most  existing personalized-aggregation-based PFL and personalized-head-based PFL methods.

\item We propose a novel PFL method, named FedAH, which, based on personalized-head-based PFL methods, performs element-wise aggregation between the local model head and the global model head to obtain an Aggregated Head, ensuring that global information is not lost in the personalized part of the model.

\item We conduct extensive experiments in the fields of computer vision (CV) and natural language processing (NLP) under two widely used scenarios. The results validate that our proposed FedAH outperforms SOTA FL methods in terms of effectiveness, scalability, and stability.
\end{itemize}

The remained paper is organized as following. Section \ref{sec:related_work} summarizes the related work. Section \ref{sec:method} demonstrates the methods used. Section \ref{sec:experiment} explains the experiments. Section \ref{sec:conclusion} concludes the paper.

\section{RELATED WORK}\label{sec:related_work}
\subsection{Personalized Federated Learning}
Traditional Federated Learning (FL) methods perform distributed machine learning through iterative communication and computation between a server and multiple clients \cite{zhang2024eliminating}. Due to the statistical heterogeneity problem in FL, a single global model often struggles to adapt well to different clients \cite{kairouz2021advances, huang2021personalized, t2020personalized}. Unlike traditional FL, Personalized Federated Learning (PFL) not only learns a global model on the server but also learns personalized models (or modules) on the clients, which has garnered significant attention for addressing the statistical heterogeneity issue in FL \cite{kairouz2021advances}. In this paper, we categorize PFL methods into the following four types:

(1) \textit{Meta-learning-based PFL}: Per-FedAvg \cite{fallah2020personalized} and FedMeta \cite{chen2018federated} combine meta-learning frameworks, leveraging the average aggregation trend of model updates to learn a global model, and obtaining personalized models by locally fine-tuning the global model on each client. However, this strategy makes it challenging for Per-FedAvg and FedMeta to find a consistent learning trend through averaging in statistically heterogeneous scenarios \cite{zhang2024eliminating}.

(2) \textit{Regularization-based PFL}: FedProx \cite{li2020federated} regularizes the difference between local model parameters and global model parameters during local model training on clients, while pFedMe \cite{t2020personalized} and Ditto \cite{li2021ditto} learn additional personalized models for each client and use proximal terms for the personalized models. Nevertheless, FedProx still learns a single global model, while pFedMe and Ditto require more memory and computational resources to store and train additional personalized models.

(3) \textit{Personalized-aggregation-based PFL}: FedFomo \cite{zhang2020personalized} and APPLE \cite{luo2022adapt} initialize local models by aggregating models of other clients locally on each client. FedAMP \cite{huang2021personalized} generates aggregated models for individual clients through an attention-inducing function and personalized aggregation. FedALA \cite{zhang2023fedala} adaptively aggregates global and local models based on the local data of each client, achieving finer-grained element-level model aggregation to initialize local models before each training iteration. However, FedALA still has room for improvement as it does not split the model into a feature extractor and a head, FedFomo and APPLE require more communication overhead, and the model-level personalized aggregation of FedAMP is not precise enough.

(4) \textit{Personalized-head-based PFL}: FedPer \cite{arivazhagan2019federated} and FedRep \cite{collins2021exploiting} learn a global feature extractor and a client-specific head, with the former training the head locally using the feature extractor and the latter locally fine-tuning the head before each training iteration of the feature extractor. However, the lack of head sharing in FedPer and FedRep leads to the loss of general information in the head, which affects the final performance of the personalized model.

Our proposed FedAH combines strategies from the third and fourth categories. Similar to FedRep, it splits the given backbone into a global feature extractor and client-specific heads, fine-tuning the heads before each training iteration of the feature extractor. Unlike FedRep, it fine-tunes the Aggregated Head, which is the element-wise aggregation of the local model head and the global model head, instead of the local model head from the previous iteration. The Aggregated Head adopts the personalized aggregation strategy of the third category, improving model performance by learning general information while achieving personalization.

\section{METHODOLOGY}\label{sec:method}
In this section, we first provide an overview of the local learning process of FedAH, then state the objectives of FL optimization, and finally perform a theoretical derivation of FedAH.

\subsection{Overview of FedAH on the client}
\begin{figure}[htpb]
    \centering
    \includegraphics[width=\linewidth]{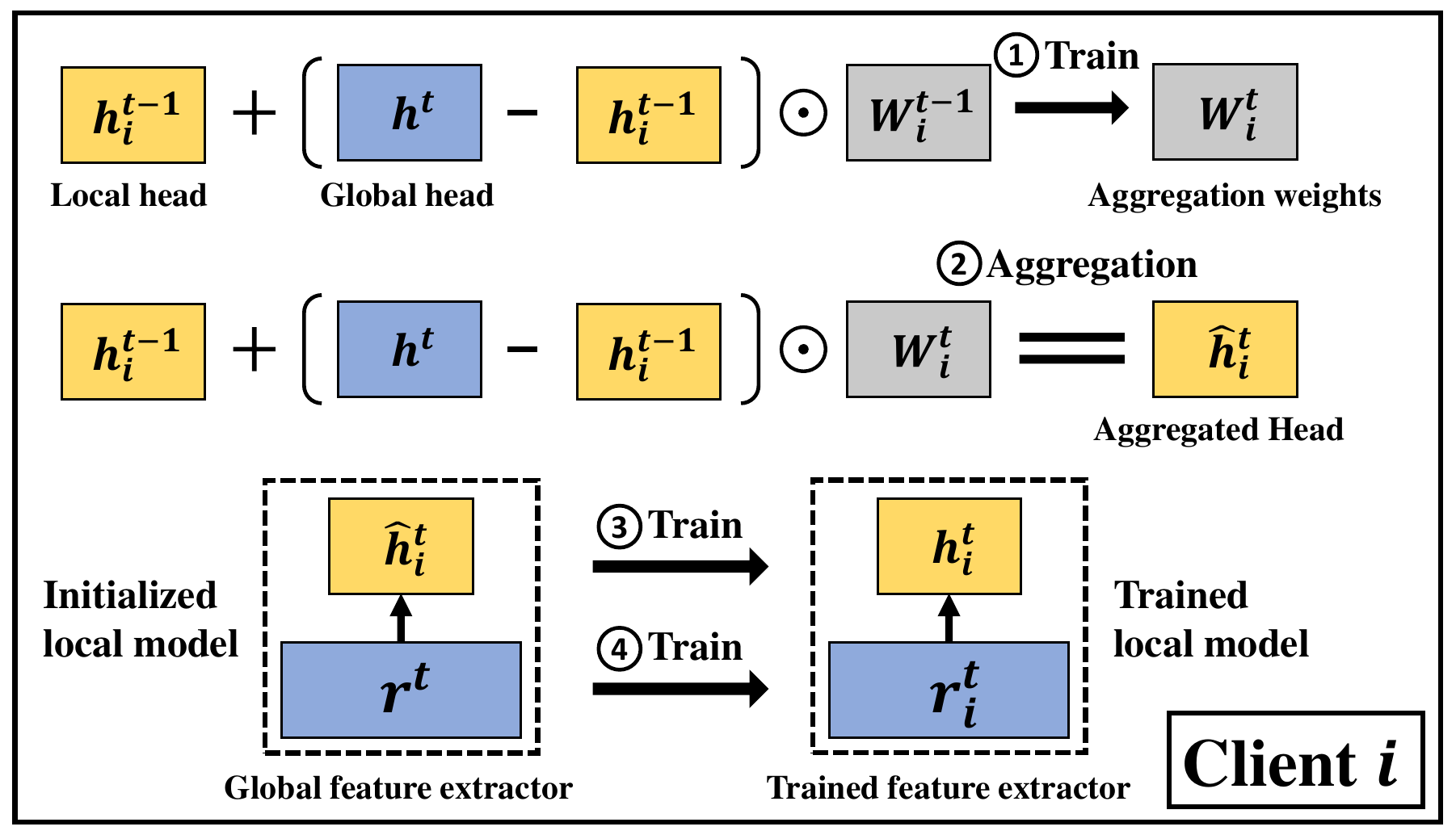}
    \caption{Local learning process of FedAH on client $i$ in the $t$-th iteration.}
    \label{fig:fedah_client}
\end{figure}

Figure \ref{fig:fedah_client} illustrates the local learning process of the proposed FedAH method on the client, which can be divided into four steps: (1) The client splits the global model downloaded from the server into a global feature extractor and a global head, and trains the aggregation weights for the head by freezing the global head and the local head from the previous iteration. (2) The client uses the new aggregation weights to perform element-wise aggregation of the global head and the local head from the previous iteration to obtain the Aggregated Head. (3) The client freezes the global feature extractor parameters and trains the Aggregated Head to get the local head for the current iteration. (4) The client freezes the local head parameters and trains the feature extractor parameters. Finally, the client obtains the trained local model and uploads it to the server, concluding the local learning process.

\subsection{Problem Statement}
In the process of FL under statistically heterogeneous settings, suppose there are $N$ clients, each with its own Non-IID and unbalanced dataset $D_1,...,D_N$. Specifically, $D_1,...,D_N$ are sampled from $N$ different distributions, and the data volumes are different. The overall objective of PFL is to collaboratively learn independent personalized models $\hat{\Theta}_1,...,\hat{\Theta}_N$ for each client under the coordination of a central server. The global loss function is minimized to obtain the reasonable personalized models:

\begin{equation}
\{\hat{\Theta}_1,...,\hat{\Theta}_N\}=\argmin_{\hat{\Theta}}
\mathcal{G}(\mathcal{L}_1,...,\mathcal{L}_N),
\end{equation}

where $\mathcal{L}_i=\mathcal{L}_i(\hat{\Theta}_i,D_i), \forall i\in [1,N]$, and $\mathcal{L}_i\left( \cdot \right) $ is the local loss function of client $i$. Typically, $\mathcal{G}(\mathcal{L}_1,...,\mathcal{L}_N)=\sum_{i=1}^N{k_i}\mathcal{L}_i $, where $k_i=|D_i|/\sum_{j=1}^N{|}D_j|$, and $|D_i|$ is the number of local data samples for client $i$.

\begin{figure}[htpb]
    \centering
    \includegraphics[width=\linewidth]{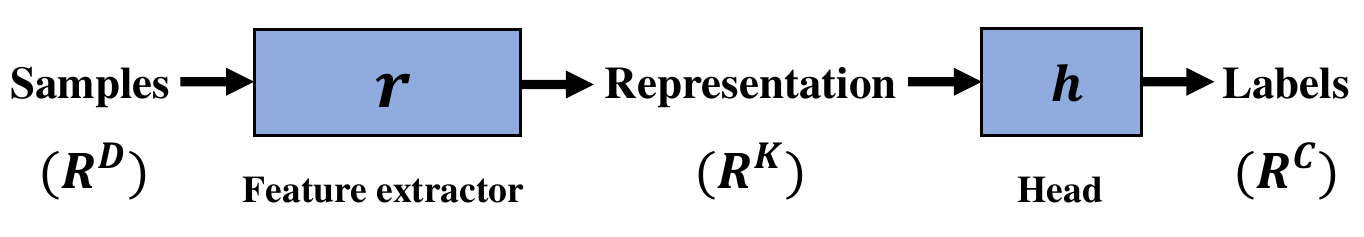}
    \caption{The inputs and outputs of the feature extractor and head in personalized-head-based PFL methods.}
    \label{fig:representation}
\end{figure}

Similar to personalized-head-based methods like FedPer and FedRep, we split the neural network model $\Theta$ into a feature extractor $\Theta_r:\mathbb{R}^D \to \mathbb{R}^K$, which maps input samples to a low-dimensional representation space, and a head $\Theta_h:\mathbb{R}^K \to \mathbb{R}^C$, which maps the representation space to the label space. Like FedRep, we treat the last fully connected (FC) layer of the neural network model as the head, and the remaining bottom layers as the feature extractor. $D$ is the dimension of the input space, $K$ is the dimension of the representation space, and $C$ is the dimension of the label space, typically $K \ll D$. As shown in Figure \ref{fig:representation}, after splitting the model into the feature extractor and head, the input sample is processed by the feature extractor to extract a low-dimensional feature representation, which is then passed through the head to obtain the label. %This process maps the high-dimensional sample space $\mathbb{R}^D$ to the low-dimensional representation space $\mathbb{R}^K$, and then to the even lower-dimensional label space $\mathbb{R}^C$.

Thus, the local model ${\Theta_i}$ of client $i$ can be transformed into a combination of its feature extractor $\Theta_{i,r}$ and head $\Theta_{i,h}$, i.e., $\Theta_i:=\{\Theta_{i,r},\Theta_{i,h}\}$. For simplicity, it is transformed to $\Theta_i:=\{r_i,h_i\}$, where $r_i$ and $h_i$ represent the parameters of the feature extractor $\Theta_r$ and the head $\Theta_h$ of client $i$, respectively.

\subsection{Aggregated Head for Personalized Federated Learning (FedAH)}
In traditional FL (e.g., FedAvg), during iteration $t$, the server randomly selects a subset $S^t$ of $N$ clients for training and aggregates all local models $\Theta _{i}^t,i \in S^t$ to obtain the global model $\Theta ^{t}$. Formally, $\Theta ^{t}$ can be derived through:

\begin{equation}
\Theta^t \leftarrow \sum_{i \in S^t} k_i \Theta_i^t.
\label{eq:get_global_model}
\end{equation}

Then, the server sends the global model $\Theta ^{t}$ to client $i$, which overwrites the local model $\Theta _{i}^{t-1}$ from the previous iteration, resulting in the initialized local model $\hat{\Theta}_{i}^{t}$ for local training, i.e., $\Theta ^{t}:=\hat{\Theta}_{i}^{t}$. However, for FedPer and FedRep, the server sends the global feature extractor $r^t$ to client $i$ to overwrite it, while the head $h^{t-1}_i$ from the local model of the previous iteration is retained, i.e., $\{r^{t}_i,h^t_i\}:=\{r^{t},h^{t-1}_i\}$. For FedAH, instead of simply retaining the local head $h^{t-1}_i$, we perform element-wise aggregation between the local head $h^{t-1}_i$ from the previous iteration and the global head $h^t$ from the current iteration to obtain the Aggregated Head $\hat{h}^t_i$. Formally:

\begin{equation}
\hat{h}_{i}^{t}:=h_i^{t-1}+(h^t-h_i^{t-1})\odot W^{t}_i,
\label{eq:get_aggregated_h}
\end{equation}

where $\odot$ denotes the Hadamard product, representing the element-wise multiplication of two matrices, and $W^{t}_i$ denotes the aggregation weights for the model head of client $i$, where $w \in [0,1],\, \forall w \in W^{t}_i$. For FedAvg, the elements of $W^{t}_i$ are all one, while for FedPer and FedRep, the elements of $W^{t}_i$ are all zero.

Client $i$ trains $W^{t}_i$ using a gradient-based learning method, initializing each element of $W^{t}_i$ with one and continuously learning the new $W^{t}_i$ based on the former $W^{t-1}_i$. Formally,

\begin{equation}
W_i^t \leftarrow W_i^{t-1} - \eta \nabla_{W_i} {\mathcal{L}_i}(\{r^{t},\hat{h}^t_i\}, D_i),
\label{eq:train_weight}
\end{equation}

where $\eta$ is the learning rate for weight learning, and all other trainable parameters, including $r^{t}, h^t$, and $h^{t-1}_i$, are frozen during each iteration. The goal of updating the weights $W^{t}_i$ is to obtain a better Aggregated Head $\hat{h}^t_i$. Additionally, to ensure $w \in [0,1],\, \forall w \in W^{t}_i$ during gradient descent, element-wise weight clipping $\sigma(w) = \max(0, \min(1, w))$ is used for regularization \cite{zhang2023fedala}.

Next,  the feature extractor parameters of the model are frozen, and the Aggregated Head $\hat{h}^t_i$ is trained to obtain the local head $h^{t}_i$ for this iteration:

\begin{equation}
h_i^t \leftarrow \hat{h}_i^{t} - \alpha \nabla_{\hat{h}_i} {\mathcal{L}_i}(\{r^{t},\hat{h}^t_i\}, D_i).
\label{eq:train_h}
\end{equation}

Then, the local head parameters are frozen, and the feature extractor parameters are trained:

\begin{equation}
r_i^t \leftarrow r^{t} - \alpha \nabla_{r} {\mathcal{L}_i}(\{r^{t},{h^t_i}\}, D_i).
\label{eq:train_r}
\end{equation}

Finally, clients upload the trained local models $\{r_i^t,h_i^t\}, \forall i \in S^t$ to the server for the next iteration of aggregation. Algorithm \ref{alg:fedah} describes the entire process of FedAH.

\begin{algorithm}[t!]
\caption{FedAH}
\label{alg:fedah}
\KwIn{$N$ clients, client joining ratio $\rho$, local loss function $\mathcal{L}_i$ and dataset $D_i$ of clien $i$, initial global model $\{r^0,h^0\}$, local learning rate $\eta$, weight learning rate $\alpha$.}
\KwOut{Personalized models $\{r_1,\hat{h}_1\}, \ldots, \{r_N,\hat{h}_N\}$}

Server sends model $\{r^0,h^0\}$ to all clients to initialize local models.\\
\For {\textnormal{iteration} $t = 1, \ldots, T$}
{Server samples a subset $S^t$ of clients based on $\rho$.\\
Server sends model $\{r^t,h^t\}$ to all client $i \in S^t$.\\
    \For {\textnormal{Client} $i \in S^t$ \textnormal{in parallel}}{
        Client $i$ trains $W^t_i$ by Equation (\ref{eq:train_weight}).  \\
        Client $i$ obtains $\hat{h}_{i}^{t}$ by Equation (\ref{eq:get_aggregated_h}).  \\
        Client $i$ trains  $h_i^t$ by Equation (\ref{eq:train_h}).  \\
        Client $i$ trains $r_{i}^{t}$ by Equation (\ref{eq:train_r}).  \\
        Client $i$ sends $\{r_i^t,h_i^t\}$ to the server.  \\
    }

Server obtains $\{r^{t+1},h^{t+1}\}$ by $\{r^{t+1},h^{t+1}\} \leftarrow \sum_{i \in S^t} k_i \{r_i^{t},h_i^{t}\}$.\\

}
\textbf{return} $\{r_1,\hat{h}_1\}, \ldots, \{r_N,\hat{h}_N\}$

\end{algorithm}

\section{Experiments}\label{sec:experiment}
\subsection{Experimental Setup}
In this section, FedAH is evaluated on various image/text classification tasks and compared with ten state-of-the-art (SOTA) FL methods including FedAvg \cite{mcmahan2017communication}, FedProx \cite{li2020federated}, Per-FedAvg \cite{fallah2020personalized}, pFedMe \cite{t2020personalized}, FedAMP \cite{huang2021personalized}, Ditto \cite{li2021ditto}, FedPer \cite{arivazhagan2019federated}, FedRep \cite{collins2021exploiting}, FedFomo \cite{zhang2020personalized}, and FedALA \cite{zhang2023fedala}. For image classification tasks, four datasets are used: MNIST \cite{lecun1998gradient}, Cifar10 \cite{krizhevsky2009learning}, Cifar100 \cite{krizhevsky2009learning}, and Tiny-ImageNet \cite{chrabaszcz2017downsampled} (100K images, 200 classes). A 4-layer CNN \cite{mcmahan2017communication} is used as the model, with an additional ResNet-18 \cite{he2016deep} for Tiny-ImageNet. The local learning rate $\eta$ is set to 0.005 for the 4-layer CNN and 0.01 for ResNet-18. For text classification tasks, the AG News \cite{zhang2015character} dataset with fastText \cite{joulin2016bag} is used, and the local learning rate for fastText is set to $\eta = 0.01$, with other settings the same as image classification tasks.

The experiments follow the FedAvg methodology, setting the batch size to 10 and the number of local model training epochs to 1. All tasks are run for 2000 iterations until all methods empirically converge. Following the methods of pFedMe and FedFomo, the total number of clients is set to 20, with a client joining ratio $\rho = 1$. The evaluation metrics used are the same as those in pFedMe, where traditional FL uses the test accuracy of the best single global model, and PFL uses the average test accuracy of the best local models. To simulate real PFL scenarios, the learned models are evaluated on the clients. 25\% of the client's local data is used as the test dataset, and the remaining 75\% is used as the training dataset. To avoid randomness, all experiments are run five times, and the mean and standard deviation are derived.

The experiments adopt two widely used scenarios to simulate heterogeneous settings. The first is the pathological heterogeneous setting \cite{mcmahan2017communication, shamsian2021personalized}, where 2/2/10 classes are sampled from a total of 10/10/100 classes for MNIST/Cifar10/Cifar100, respectively, with non-overlapping data samples. Specifically, similar to FedAvg, clients are grouped with the same labels but with imbalanced data. The second scenario is the practical heterogeneous setting \cite{lin2020ensemble, li2021model}, controlled by a Dirichlet distribution denoted as $Dir(\beta)$. The smaller the $\beta$, the greater the heterogeneity of the environment is. In the experiments, $\beta = 0.1$ is set as the default heterogeneity setting \cite{lin2020ensemble, zhang2023fedala}.

In the experiments, our proposed FedAH is implemented using PyTorch-1.12.1 and simulate FL on a server equipped with an AMD Epyc 7302 16-core processor x 64, 8 NVIDIA GeForce RTX 3090 GPUs, 472.2GB of memory, and running Ubuntu 20.04.5 operating system.

\subsection{Effectiveness}

\begin{table*}[htpb] % The table* environment makes the table span both columns
\centering % Center the table
\caption{The test accuracy (\%) of the image/text classification tasks in the pathological/practical heterogeneous setting}

\begin{tabular}{l|lll|llllll}
\toprule
Settings  & \multicolumn{3}{c|}{Pathological heterogeneous setting}                                                                      & \multicolumn{5}{c}{Practical heterogeneous setting ($\beta = 0.1$)}                                                                   \\
\midrule
Datasets    & \multicolumn{1}{c}{MNIST} & \multicolumn{1}{c}{Cifar10} & \multicolumn{1}{c|}{Cifar100} & \multicolumn{1}{c}{MNIST} & \multicolumn{1}{c}{Cifar10} & \multicolumn{1}{c}{Cifar100} & \multicolumn{1}{c}{TINY} & \multicolumn{1}{c}{TINY*} & \multicolumn{1}{c}{AG News}\\
\midrule
FedAvg     & 97.93±0.05 & 55.09±0.83 & 25.98±0.13 & 98.81±0.01 & 59.16±0.47 & 31.89±0.47 & 19.46±0.20 & 19.45±0.13 & 79.57±0.17 \\
FedProx    & 98.01±0.09 & 55.06±0.75 & 25.94±0.16 & 98.82±0.01 & 59.21±0.40 & 31.99±0.41 & 19.37±0.22 & 19.27±0.23 & 79.35±0.23 \\
\midrule
Per-FedAvg & 99.63±0.02 & 89.63±0.23 & 56.80±0.26 & 98.90±0.05 & 87.74±0.19 & 44.28±0.33 & 25.07±0.07 & 21.81±0.54 & 93.27±0.25 \\
pFedMe     & 99.75±0.02 & 90.11±0.10 & 58.20±0.14  & 99.52±0.02 & 88.09±0.32 & 47.34±0.46 & 26.93±0.19 & 33.44±0.33 & 91.41±0.22 \\
FedAMP     & 99.76±0.02 & 90.79±0.16 & 64.34±0.37 & 99.47±0.02 & 88.70±0.18 & 47.69±0.49 & 27.99±0.11 & 29.11±0.15 & 94.18±0.09 \\
Ditto      & 99.81±0.00 & 92.39±0.06 & 67.23±0.07 & 99.64±0.00 & 90.59±0.01 & 52.87±0.64 & 32.15±0.04 & 35.92±0.43 & 95.45±0.17 \\
FedPer     & 99.70±0.02 & 91.15±0.21 & 63.53±0.21 & 99.47±0.04 & 89.22±0.33 & 49.63±0.54 & 33.84±0.34 & 38.45±0.85 & 95.54±0.32 \\
FedRep     & 99.77±0.03 & 91.93±0.14 & 67.56±0.31 & 99.48±0.02 & 90.40±0.24 & 52.39±0.35 & 37.27±0.20 & 39.95±0.61 & 96.28±0.14 \\
FedFomo    & 99.83±0.00 & 91.85±0.02 & 62.49±0.22 & 99.33±0.04 & 88.06±0.02 & 45.39±0.45 & 26.33±0.22 & 26.84±0.11 & 95.84±0.15 \\
FedALA     & 99.88±0.01 & 92.44±0.02 & 67.83±0.06 & \textbf{99.71±0.00} & 90.67±0.03 & 55.92±0.03 & 40.54±0.02 & 41.94±0.05 & 96.52±0.08 \\               
\midrule
FedAH & \textbf{99.90±0.01} & \textbf{92.57±0.05} & \textbf{69.70±0.09} & 99.66±0.00 & \textbf{91.11±0.04} & \textbf{58.79±0.05} & \textbf{42.90±0.03} & \textbf{44.30±0.14} & \textbf{96.65±0.11} \\  
\bottomrule
\end{tabular}
\label{tab:effectiveness}
\end{table*}

The experiments denote “TINY” and “TINY*” to represent the use of a 4-layer CNN and ResNet-18 on Tiny-ImageNet, respectively. In our experiments, the learning rate of the aggregation weights in FedAH is set to be the same as the local learning rate, with the number of training epochs per iteration set to 1. Table \ref{tab:effectiveness} shows that, except for the MNIST dataset under the default practical heterogeneous setting, FedAH outperforms all FL methods in terms of test accuracy across five benchmark datasets in CV and NLP, particularly on larger datasets (Cifar100 and TINY) and with more complex models (ResNet-18). On the Cifar100 dataset under the default practical heterogeneous setting, FedAH exceeds the second-best method, FedALA, by 2.87\%. The poor performance of FedAvg in Table \ref{tab:effectiveness} is evident, as a single global model trained by traditional FL methods cannot fit well to the local data of all clients in a heterogeneous setting. Next, we analyze the reasons that FedAH outperforms the other four categories of PFL methods.

\emph{Meta-learning-based PFL}. Compared to traditional FL methods, PFL methods generally perform better. However, among these PFL methods, Per-FedAvg has the lowest test accuracy because it only obtains an initial global model that corresponds to the learning trend of all clients, making it difficult to meet the trends of each personalized model. In contrast, FedAH splits the model into a feature extractor and a head, achieving personalization through the client-specific head, which better fits the heterogeneous data of different clients, thus performing better.

\emph{Regularization-based PFL}. FedProx performs similarly to FedAvg because it still learns a single global model. Both pFedMe and Ditto use proximal terms to learn additional personalized models, but pFedMe learns from the local model while Ditto learns from the global model. Since Ditto can extract global information from the global model, its performance is better than that of pFedMe. However, using proximal terms to learn personalized models is an implicit method, and its effect is not as good as the explicit method of FedAH, which splits and aggregates the head.

\emph{Personalized-aggregation-based PFL}. The model-level personalized aggregation of FedFomo and FedAMP is not precise enough and may introduce useless information from the global model into the local model. Additionally, FedFomo requires downloading multiple other clients' models in each iteration, resulting in higher communication overhead. FedALA, by adaptively learning aggregation weights, can accurately capture the required information from the global model, thus outperforming FedFomo and FedAMP. However, FedALA does not explicitly split the model into feature extractor and head, and alternating training of these two parts can significantly improve model performance, making FedALA perform worse than FedAH in most of the experiments.

\begin{figure}[htpb]
    \centering
    \includegraphics[width=\linewidth]{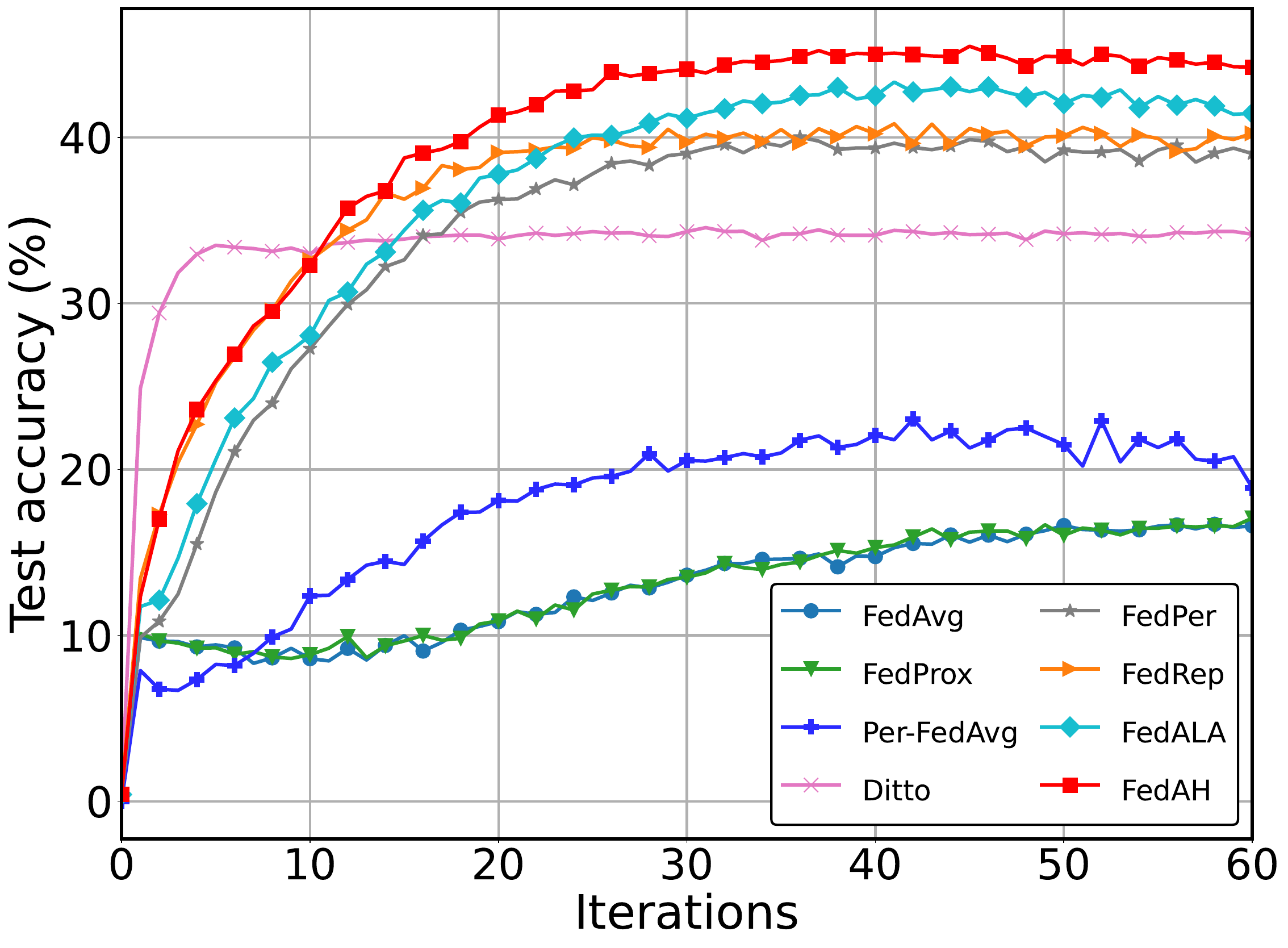}
    \caption{Test accuracy (\%) curves of different methods on Tiny-ImageNet using ResNet-18.}
    \label{fig:acc_tiny2}
\end{figure}

\emph{Personalized-head-based PFL}. Although FedPer and FedRep split the model like FedAH, they only share the feature extractor but not the head, losing the global information of the model head. FedAH aggregates the local model head and the global model head at the element level to obtain the Aggregated Head. It thereby introduces global information from the global model head,and improves the overall performance of the model, thus performing better.

\begin{table*}[htpb] % The table* environment makes the table span both columns
\centering % Center the table
\caption{The test accuracy (\%) of the image/text classification tasks for heterogeneity and scalability.}

\begin{tabular}{l|lll|lllll}
\toprule
& \multicolumn{3}{c|}{Heterogeneity}                             & \multicolumn{5}{c}{Scalability}     \\       
\midrule
Datasets    & \multicolumn{2}{c}{TINY} & \multicolumn{1}{c|}{AG News}  & \multicolumn{5}{c}{Cifar100}\\
\midrule
Settings    & \multicolumn{1}{c}{$\beta = 0.01$} & \multicolumn{1}{c}{$\beta = 0.5$} & \multicolumn{1}{c|}{$\beta = 1$} & \multicolumn{1}{c}{$N = 10$} & \multicolumn{1}{c}{$N = 30$} & \multicolumn{1}{c}{$N = 50$} & \multicolumn{1}{c}{$N = 100$} & \multicolumn{1}{c}{$N = 200$} \\
\midrule
FedAvg     & 15.70±0.46 & 21.14±0.47 & 87.12±0.19 & 31.47±0.01 & 31.15±0.05 & 31.90±0.27 & 31.95±0.37 & 31.20±0.58 \\
FedProx    & 15.66±0.36 & 21.22±0.47 & 87.21±0.13 & 31.24±0.08 & 31.21±0.08 & 31.94±0.30 & 31.97±0.24 & 31.22±0.62 \\
\midrule
Per-FedAvg & 39.39±0.30 & 16.36±0.13 & 87.08±0.26 & 37.24±0.12 & 41.57±0.21 & 44.31±0.20 & 36.07±0.24 & \multicolumn{1}{c}{—}         \\
pFedMe     & 41.45±0.14 & 17.48±0.61 & 87.08±0.18 & 44.06±0.29 & 47.04±0.28 & 48.36±0.64 & 46.45±0.18 & 39.55±0.61 \\
FedAMP     & 48.42±0.06 & 12.48±0.21 & 83.35±0.05 & 49.23±0.18 & 45.33±0.04 & 44.39±0.35 & 40.43±0.17 & 35.40±0.70 \\
Ditto      & 50.62±0.02 & 18.98±0.05 & 91.89±0.17 & 52.32±0.19 & 52.53±0.42 & 54.22±0.04 & 52.89±0.22 & 35.18±0.53 \\
FedPer     & 51.83±0.22 & 17.31±0.19 & 91.85±0.24 & 50.31±0.19 & 44.98±0.20 & 44.22±0.18 & 40.37±0.41 & 34.99±0.48 \\
FedRep     & 55.43±0.15 & 16.74±0.09 & 92.25±0.20 & 52.89±0.10 & 50.24±0.01 & 47.41±0.18 & 44.61±0.20 & 36.79±0.60 \\
FedFomo    & 46.36±0.54 & 11.59±0.11 & 91.20±0.18 & 46.71±0.23 & 43.20±0.05 & 42.56±0.33 & 38.91±0.08 & 34.79±0.71 \\
FedALA & 55.75±0.02          & 27.85±0.06          & 92.45±0.10          & 56.31±0.09          & 56.01±0.13          & 55.61±0.02          & 54.68±0.57          & 45.78±0.83          \\
\midrule
FedAH  & \textbf{56.14±0.07} & \textbf{27.91±0.04} & \textbf{92.59±0.08} & \textbf{58.44±0.03} & \textbf{58.13±0.10} & \textbf{57.53±0.15} & \textbf{56.05±0.26} & \textbf{48.37±0.65}          \\
\bottomrule
\end{tabular}
\label{tab:scalability}
\end{table*}

Overall, by adaptively learning the aggregation weights of the head, FedAH can accurately capture the required global information in the global head and utilize the Aggregated Head with global information introduced as the initialized local head. This addresses the shortcomings of FedPer and FedRep. Additionally, FedAH follows the training methods of FedPer and FedRep, alternating the training of the head and the feature extractor, which is more effective than the way of directly training the entire model. Therefore, by combining the advantages of personalized-aggregation-based and personalized-head-based PFL methods, FedAH performs the best among all the SOTA methods.

Figure \ref{fig:acc_tiny2} shows the test accuracy curves of FedAH and various FL methods on TINY* under the default heterogeneous setting. FedProx, which adds a proximal term to FedAvg, has the minimal effect, and their accuracy curves are quite close. Both methods converge slowly and require more iterations due to their inability to train personalized models. Per-FedAvg performs the worst among the PFL methods, and its accuracy curve shows a noticeable decline after reaching the peak. This is because its local fine-tuning strategy leads to severe overfitting in the later iterations. Ditto converges quickly and remains stable after convergence, but its peak accuracy is amid. Typical personalized-head-based methods like FedPer and FedRep, perform well and are very close in performance. However, their performance in later stages is not commensurate with FedAH because their model heads are intractable of global knowledge. By combining the advantages of FedRep and FedALA, FedAH outperforms both and achieves the best performance among all methods. Additionally, FedAH maintains its performance well after convergence because the Aggregated Head introduces global information, which alleviates overfitting in the personalized models.

\subsection{Different Heterogeneity}

To verify the effectiveness of FedAH under different degrees of heterogeneity settings, experiments are conducted on the Tiny-ImageNet and AG News datasets by changing the $\beta$ of $Dir(\beta)$. The smaller the $\beta$, the greater the heterogeneity of the settings is. As shown in Table \ref{tab:scalability}, the test accuracy of FedAH remains superior to all methods. Most PFL methods perform better in settings with stronger heterogeneity, while their test accuracy drops significantly when the heterogeneity is weaker, i.e., with larger $\beta$. When the degree of heterogeneity in Tiny-ImageNet reaches $\beta = 0.5$, only FedALA and FedAH have test accuracy higher than the traditional FedAvg, as they can accurately capture global information from the global model/head through adaptive model/head aggregation during local learning, thus maintaining excellent performance in settings with weaker heterogeneity.

\subsection{Scalability}

To verify the scalability of FedAH, five experiments are conducted following the methodology of MOON \cite{li2021model} on the Cifar100 dataset under the default heterogeneous setting, with the number of clients $N$ set to $N = 10, 30, 50, 100, \text{and} \; 200$. Since the total data volume on the Cifar100 dataset is fixed, the local data volume (average) of each client decreases as the number of clients increases. As shown in Table \ref{tab:scalability}, when the number of clients increases to 100 and 200, the test accuracy of most PFL methods drops significantly due to the lack of local data on clients, while the test accuracy of FedAH remains superior to all other methods. This is because, in the case of sparse local data, it is more important to accurately capture global information from the collaboratively trained global model (or head), which can learn more global knowledge.

\subsection{Computation and Communication Overhead}

As shown in Table \ref{tab:overhead}, the experiments record the total time and the number of iterations required for convergence (determined by an early stopping mechanism) for each FL method, and calculate the average computation time per iteration. Per-FedAvg requires more time per iteration than most methods because it needs to fine-tune the local model. Since learning personalized models requires additional training steps, pFedMe has the highest computation overhead per iteration, and Ditto faces a similar situation. FedAH requires more training time per iteration, only less than the aforementioned three methods, because FedAH needs to additionally train the aggregation weights of the head and fine-tune the Aggregated Head in each iteration. However, using the Aggregated Head as a better initial head allows the model to converge quickly, resulting in relatively low total computation time and number of iterations.

As shown in Table \ref{tab:overhead}, we can theoretically compare the communication overhead of each FL method in a single iteration for one client. Most methods only need to upload and download the model once per iteration, so with the same number of model parameters and iterations, their communication overhead is the same. FedPer and FedRep transmit only the feature extractor part in each iteration, resulting in the lowest communication overhead per iteration. FedFomo requires downloading multiple other client models in each iteration, leading to higher communication overhead. FedAH has lower communication overhead since it converges fast, requiring fewer iterations.

\begin{table}[htpb]
\centering
\caption{The computation overhead on Tiny-ImageNet using ResNet-18 and communication overhead (parameters transmitted per iteration). $\Sigma$ is the number of parameters in the backbone. $\alpha$ $(\alpha < 1)$ is the ratio of parameters of the feature extractor in the backbone. $n$ $(n \geq 1)$ is the number of other clients each client receives in FedFomo.}

\begin{tabular}{l|cc|cc}

\toprule
           & \multicolumn{2}{c|}{Computation} & \multicolumn{2}{c}{Communication} \\
\midrule
Methods    & Total time     & Time/iter.  & Iterations   & Param./iter.   \\
\midrule
FedAvg     & 352 min	& 2.13 min	&165   & $2\ast\Sigma$   \\
FedProx    & 316 min	& 2.41 min	&131   & $2\ast\Sigma$  \\
\midrule
Per-FedAvg & 260 min	&4.41 min	&59      & $2\ast\Sigma$     \\
pFedMe     & 1757 min	&8.49 min	&207         & $2\ast\Sigma$    \\
FedAMP     & 82 min	    &2.28 min	&36    & $2\ast\Sigma$  \\
Ditto     & 104 min	    &4.53 min	&23         & $2\ast\Sigma$    \\
FedPer     & 198 min	&2.15 min	&92       & $2\ast\alpha\ast\Sigma$  \\
FedRep    & 281 min     &2.46 min	&114       & $2\ast\alpha\ast\Sigma$  \\
FedFomo   & 170 min  	&2.27 min	&75     &  $(1+n)\ast\Sigma$   \\
FedALA   & 130 min	   &2.45 min	&53         & $2\ast\Sigma$   \\
\midrule
FedAH    & 121 min	   &3.03 min	&40        & $2\ast\Sigma$    \\
\bottomrule
\end{tabular}
\label{tab:overhead}
% \vspace{-0.1in}
\end{table}

\subsection{Stability}

In real-world scenarios, some clients may unexpectedly drop out in a certain iteration and rejoin in a subsequent iteration due to reasons like insufficient battery power, lack of computing and storage resources, or network instability. To compare the performance of different PFL methods under such conditions, we simulate this scenario by changing the client joining ratio $\rho$ in each iteration on the Cifar100 dataset. Specifically, instead of fixing the $\rho$ value,  $\rho$ values are uniformly sampled within a given range in each iteration. A larger range of $\rho$ indicates a more unstable scenario. Compared to the settings of other FL methods with a fixed client joining ratio, our experiment is significantly closer to real-world scenarios. As shown in Table \ref{tab:stability}, with the increase in the range of $\rho$, i.e., the more frequent the random dropout and joining behavior of clients happen, leading to the decrease of  the mean and standard deviation of test accuracy for most methods. Some PFL methods, such as pFedMe and Ditto, perform much worse with a larger range of $\rho$. Compared to $\rho = 1$, their test accuracy decrease by 6.65\% and 2.26\%, respectively, when $\rho \in [0.1,1]$. For $\rho$ in the same range,  the standard deviations of Per-FedAvg, pFedMe, and Ditto are all greater than 1\%, indicating their unstable performance in dynamic scenarios. However, The test accuracy of FedAH remains superior to all methods in such dynamic scenarios, with only a slight increase in standard deviation, demonstrating its stability. This is because clients joining FedAH train the aggregation weights of the head at the beginning of each iteration, allowing the Aggregated Head to quickly adapt to the changing environment. Thus, FedAH maintains its advantage and stable performance in these dynamic scenarios.

\begin{table}[htpb]
\centering % Center the table
\caption{The test accuracy (\%) of the PFL methods on Cifar100 ($N = 50$, $\beta = 0.1$ and $\rho \leq 1$) when clients unexpectedly drop out.}

\begin{tabular}{l|lll}
\toprule
Ratios     & $\rho = 1$ & $\rho \in [0.5,1]$ & $\rho \in [0.1,1]$ \\
\midrule
Per-FedAvg & 44.31±0.20 & 43.66±1.38         & 43.63±1.07         \\
pFedMe     & 48.36±0.64 & 43.28±0.85         & 41.71±1.02         \\
FedAMP     & 44.39±0.35 & 42.91±0.08         & 42.92±0.14         \\
Ditto      & 50.59±0.22 & 49.78±0.36         & 48.33±3.27         \\
FedPer     & 44.22±0.18 & 44.12±0.21         & 44.07±0.27         \\
FedRep     & 47.41±0.18 & 46.93±0.21         & 46.61±0.22         \\
FedFomo    & 42.56±0.33 & 40.96±0.02         & 40.93±0.07         \\
FedALA     & 55.61±0.02 & 55.14±0.05        & 54.78±0.14   \\
\midrule
FedAH  & \textbf{57.53±0.15} & \textbf{57.35±0.23} & \textbf{56.92±0.22}  \\
\bottomrule
\end{tabular}
\label{tab:stability}
\end{table}

\begin{table}[htpb]
\centering % Center the table
\caption{The test accuracy (\%) of the PFL methods on Cifar10 ($N = 20$, $\beta = 0.1$) with different local epochs.}

\begin{tabular}{l|llll}
\toprule
Local epochs     & \multicolumn{1}{c}{5} & \multicolumn{1}{c}{10} & \multicolumn{1}{c}{20} & \multicolumn{1}{c}{40} \\
\midrule
FedAvg     & 57.51±0.35 & 57.55±0.32 & 57.28±0.23 & 56.27±0.29 \\
FedProx    & 57.48±0.28 & 57.69±0.31 & 57.53±0.33 & 56.18±0.24 \\
\midrule
Per-FedAvg & 86.13±0.12 & 86.09±0.19 & 85.57±0.15 & 85.45±0.16 \\
pFedMe     & 88.72±0.02 & 88.58±0.17 & 88.37±0.14 & 88.16±0.20 \\
FedAMP     & 88.72±0.21 & 88.77±0.27 & 88.76±0.30 & 88.70±0.26 \\
Ditto      & 90.79±0.21 & 90.59±0.06 & 90.34±0.23 & 90.02±0.38 \\
FedPer     & 89.62±0.12 & 89.73±0.31 & 89.79±0.35 & 89.49±0.55 \\
FedRep     & 90.20±0.41 & 90.08±0.26 & 89.46±0.13 & 89.22±0.25 \\
FedFomo    & 88.39±0.15 & 88.43±0.16 & 88.41±0.13 & 88.13±0.32 \\
FedALA     & 90.57±0.19 & 90.41±0.21 & 90.35±0.15 & 89.93±0.27          \\
\midrule
FedAH  & \textbf{91.03±0.10} & \textbf{90.87±0.19} & \textbf{90.72±0.17} & \textbf{90.29±0.26} \\
\bottomrule
\end{tabular}
\label{tab:local_epochs}
\end{table}

\subsection{Different Local Epochs}

To verify the effectiveness of FedAH under different local epochs, four experiments are conducted on the Cifar10 dataset under the default heterogeneous setting, with local epochs set to 5, 10, 20, and 40 while keeping other conditions unchanged. For most FL methods, increasing local epochs can reduce the total number of communication iterations but also increases the computational overhead per iteration and carries the risk of overfitting \cite{mcmahan2017communication}. As shown in Table \ref{tab:local_epochs}, the test accuracy of FedAH remains superior to all methods across different local epochs settings. In heterogeneous settings, more local training increases the disparity of models on different clients, which is detrimental to server model aggregation and prone to overfitting. Therefore, the test accuracy of most FL methods decreases with the increase in local epochs, and FedAH follows the same trend.

\section{Conclusion}\label{sec:conclusion}

In this paper, we propose Federated Learning with Aggregated Head (FedAH), a novel personalized federated learning method that addresses the loss of global information of the model head in personalized-head-based PFL methods. By performing element-level aggregation between the local model head and the global model head, FedAH introduces global knowledge into the personalized model heads, thereby enhancing the overall model performance. Our extensive experiments on five benchmark datasets in computer vision and natural language processing demonstrate that FedAH outperforms ten state-of-the-art FL methods by 2.87\% in test accuracy. Additionally, FedAH maintains its advantage under different degrees of heterogeneity, with increasing numbers of clients, and in scenarios where clients drop out unexpectedly, showcasing its effectiveness, scalability, and stability.

\bibliographystyle{IEEEtran}
\bibliography{IEEEabrv,references}

\vspace{12pt}

\end{document}